\documentclass[manuscript]{acmart}

\usepackage{microtype}
\usepackage{booktabs}
\usepackage{subfig}
\usepackage{mathtools}
\usepackage{graphicx}

\DeclareMathOperator*{\argmax}{arg\,max}

\DeclarePairedDelimiter{\abs}{\vert}{\vert}

\newcommand{\comparewidth}{3.2in}
\newcommand{\individualwidth}{2.0in}
\newcommand{\cpparetowidth}{3.35in}
\newcommand{\ibparetowidth}{3.35in}
\newcommand{\rulewidth}{2.9in}

\copyrightyear{2018} 
\acmYear{2018} 
\setcopyright{acmlicensed}
\acmConference[GECCO '18 Companion]{Genetic and Evolutionary Computation Conference Companion}{July 15--19, 2018}{Kyoto, Japan}
\acmBooktitle{GECCO '18 Companion: Genetic and Evolutionary Computation Conference Companion, July 15--19, 2018, Kyoto, Japan}
\acmPrice{15.00}
\acmDOI{10.1145/3205651.3208277}
\acmISBN{978-1-4503-5764-7/18/07}

\begin{document}
\title{Generating Interpretable Fuzzy Controllers using Particle Swarm Optimization and Genetic Programming}

\author{Daniel~Hein}
\orcid{0000-0002-8375-1592}
\additionalaffiliation{
  \institution{Technical University of Munich, Departement of Informatics}
  \streetaddress{Boltzmannstr. 3}
  \city{Garching} 
  \postcode{85748}
  \country{Germany}
}
\affiliation{
  \institution{Siemens AG, 
  Corporate Technology,}
  \streetaddress{Otto-Hahn-Ring 6}
  \city{Munich}  
  \postcode{81739}
  \country{Germany}
}
\email{hein.daniel@siemens.com}

\author{Steffen~Udluft}
\affiliation{
  \institution{Siemens AG, 
  Corporate Technology,}
  \streetaddress{Otto-Hahn-Ring 6}
  \city{Munich}  
  \postcode{81739}
  \country{Germany}
}
\email{steffen.udluft@siemens.com}

\author{Thomas~A.~Runkler}
\affiliation{
  \institution{Siemens AG, 
  Corporate Technology,}
  \streetaddress{Otto-Hahn-Ring 6}
  \city{Munich}  
  \postcode{81739}
  \country{Germany}
}
\email{thomas.runkler@siemens.com}

\renewcommand{\shortauthors}{D. Hein et al.}
\renewcommand{\shorttitle}{Generating Interpretable Fuzzy Controllers using PSO and GP}

\begin{abstract}
Autonomously training interpretable control strategies, called policies, using pre-existing plant trajectory data is of great interest in industrial applications.
Fuzzy controllers have been used in industry for decades as interpretable and efficient system controllers.
In this study, we introduce a fuzzy genetic programming (GP) approach called \textit{fuzzy GP reinforcement learning} (FGPRL) that can select the relevant state features, determine the size of the required fuzzy rule set, and automatically adjust all the controller parameters simultaneously.
Each GP individual's fitness is computed using model-based batch reinforcement learning (RL), which first trains a model using available system samples and subsequently performs Monte Carlo rollouts to predict each policy candidate's performance.
We compare FGPRL to an extended version of a related method called \textit{fuzzy particle swarm reinforcement learning} (FPSRL), which uses swarm intelligence to tune the fuzzy policy parameters.
Experiments using an industrial benchmark show that FGPRL is able to autonomously learn interpretable fuzzy policies with high control performance.
\end{abstract}

%
%



\maketitle

\section{Introduction}

In typical industrial applications, such as controlling wind or gas turbines, control strategies, known as policies, that can be interpreted and controlled by humans, are of great interest~\citep{maes:12}.
However, using domain experts to manually design such policies is complicated and sometimes infeasible, since it requires the plant's system dependencies to be modeled in great detail with dedicated mathematical representations.
Since such representations cannot be found for many real-world applications, policies have to be learned via reward samples from the plant itself.
Reinforcement learning (RL)~\citep{sutton:98} is capable of determining such policies using only the available system data.

Recently, \textit{fuzzy particle swarm reinforcement learning} (FPSRL) has been proposed, and it has been shown that an evolutionary computation method, namely particle swarm optimization (PSO), can be successfully combined with fuzzy rule-based systems to generate interpretable RL policies~\citep{hein:17c}.
This can be achieved by first training a model on a batch of pre-existing state-action trajectory samples and subsequently conducting model-based RL.
This step uses PSO to optimize a predefined set of fuzzy rule parameters.

FPSRL has been applied to several well-known RL benchmarks, such as the mountain car and cart-pole problems~\citep{hein:17c}.
While such simple benchmark problems are well-suited to introducing a new method and comparing its performance to that of standard approaches, their easy-to-model dynamics and low-dimensional state and action spaces share few similarities with real-world industrial applications.
Real applications usually have high-dimensional continuous state and action spaces.
Applying FPSRL to systems with many state features often yields non-interpretable fuzzy systems since every fuzzy rule contains all the state dimensions, including redundant or irrelevant state dimensions, in its membership function by default.

In this paper, we propose an approach to efficiently determine the most important state features with respect to the optimal policy.
Selecting only the most important features prior to policy parameter training makes the production of interpretable fuzzy policies using FPSRL possible again.

However, performing a heuristic feature selection initially and subsequently creating policy structures manually is a feasible but limited approach; in high-dimensional state and action spaces, the effort involved grows exponentially.

Instead, we propose as main contribution of our work \textit{fuzzy genetic programming reinforcement learning} (FGPRL), an approach, such as FPSRL, that is based on model-based batch RL.
By creating fuzzy rules using genetic programming (GP) rather than tuning the fuzzy rule parameters via PSO, FGPRL eliminates the manual feature selection process.
This GP technique is able to automatically select the most important features as well as the most compact fuzzy rule representation with respect to a certain level of performance.
Moreover, it returns not just one solution to the problem but a whole Pareto front containing the best-performing solutions for many different levels of complexity.

Although genetic fuzzy systems have demonstrated their ability to learn and adapt to solve different types of problems in various application domains, GP-generated fuzzy logic controllers have never been combined with a model-based batch RL approach so far.

Combining a fuzzy system's approximate reasoning with an evolutionary algorithm's ability to learn allows the proposed method to learn human-interpretable soft-computing solutions autonomously.
\citet{cordon:04} provide an extensive overview of previous genetic fuzzy rule-based systems.
While most of the existing research in this area has focused on genetic tuning of scaling and membership functions as well as genetic learning of rule and knowledge bases, less attention has been paid to using GP to design fuzzy rule-based systems.
Since GP is concerned with automatically generating computer programs~\citep{koza:92}, it should theoretically be able to both learn rule and knowledge bases as well as tune scale and membership functions simultaneously~\citep{geyer:95}.

Fuzzy rule-based systems have been combined with GP for modeling~\citep{hoffmann:01} and classification~\citep{ramos:00,sanchez:01,chien:02,berlanga:10} tasks.
In the optimal system control field considered in this paper, early applications that combine GP and fuzzy rule-based systems for mobile robot path tracking have been demonstrated~\citep{tunstel:96}.
Type-constraint GP has also been used to define fuzzy logic controller rule-bases for the cart-centering problem~\citep{alba:96,alba:99}.
Memetic GP, which combines local and global optimization, has been used to train Takagi-Sugeno fuzzy controllers to solve the cart-pole balancing problem~\citep{tsakonas:13}.
Recently, based on the GP fuzzy inference system (GPFIS), GPFIS-control has been proposed~\citep{koshiyama:14}.
They used multi-gene GP to automatically train a fuzzy logic controller and tested its performance on the cart-centering and inverted pendulum problems.

In this study, we apply FPSRL and FGPRL to two different benchmarks, namely the cart-pole swing-up and industrial benchmarks, to compare their RL policy performance and the interpretability of their fuzzy system controllers.

\section{Policy Generation Methods}
\label{section:methods}

\begin{figure}
	\centering
	\includegraphics[width=\comparewidth]{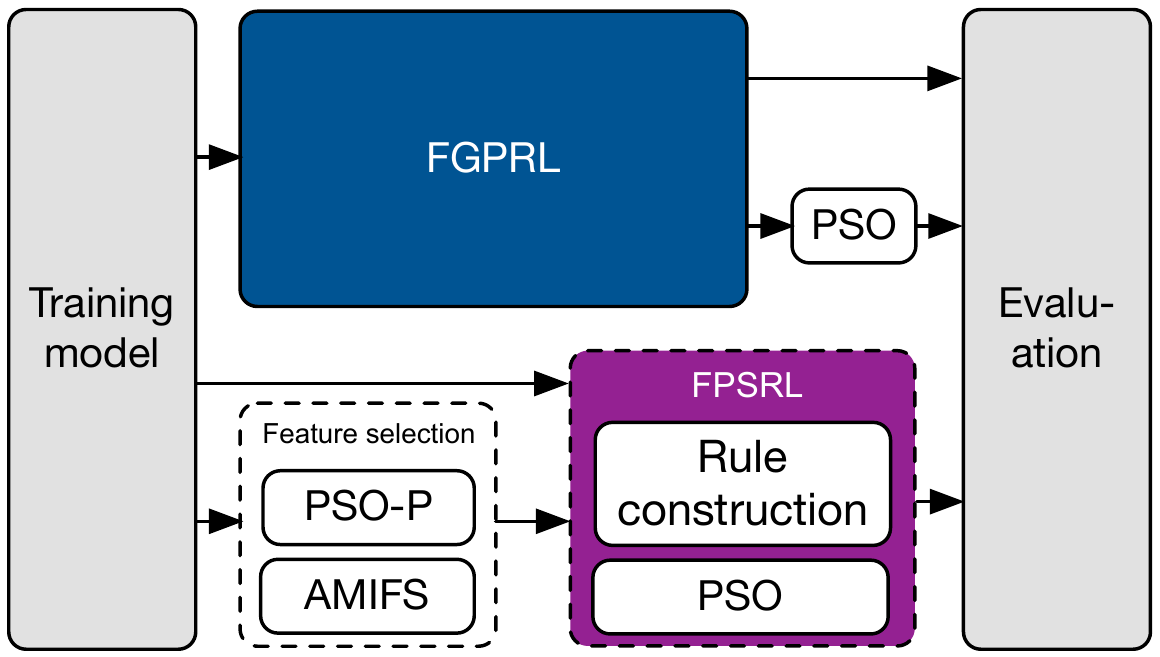}
	\caption{Comparing FPSRL to FGPRL}
	\label{figure:compare}
\end{figure}

This paper compares two approaches to generate fuzzy RL policies from a batch of previously generated state-action trajectories (Fig.~\ref{figure:compare}).
FPSRL, first proposed in~\citep{hein:17c}, tunes the fuzzy membership parameters of a predefined fuzzy set.
Here, we extend FPSRL by adding an initial feature selection step, thus enabling its application to RL problems with high-dimensional state spaces.
In addition, we compare FPSRL to a new approach, called FGPRL, that uses GP to create fuzzy RL policies using the same underlying model-based RL fitness function as FPSRL.

\subsection{Model-based Reinforcement Learning}
\label{section:rl}

Inspired by behaviorist psychology, RL is concerned with the actions software agents should take in an environment to maximize their received accumulated rewards.
In RL, the agents are not explicitly told the actions they are supposed to take; instead, they must learn the best strategy by observing the rewards given by the environment in response to their actions. 
In general, their actions can affect both the next reward and all subsequent rewards~\citep{sutton:98}.

In the RL formalism, each agent observes the system state $\mathbf{s}_t \in \mathcal S$ at each discrete time step $t=0,1,2,\ldots$ and takes an action $\mathbf{a}_t \in \mathcal A$, where $\mathcal S$ and $\mathcal A$ are the state and action spaces, respectively.
In deterministic systems, state transitions can be expressed as function $g:\mathcal S \times \mathcal A \rightarrow \mathcal S$ with $g(\mathbf{s}_t,\mathbf{a}_t)=\mathbf{s}_{t+1}$.
The corresponding rewards are given by reward function $r:\mathcal S \times \mathcal A \times \mathcal S \rightarrow \mathbb{R}$, with $r(\mathbf{s}_t,\mathbf{a}_t,\mathbf{s}_{t+1})=r_{t+1}$. 
Thus, the RL problem's optimal solution is the policy that maximizes the expected accumulated rewards.

In the proposed approach, the goal is to find the best policy $\pi\in\Pi$, with $\Pi$ being the set of all possible fuzzy RL policies.
Policies associate every state $\mathbf{s}_t$ with an action $\pi(\mathbf{s}_t)=\mathbf{a}_t$, and their performance, for a given starting state $\mathbf{s}_t$, is measured by the return $\mathcal{R}(\mathbf{s}_t,\pi)$, i.e., the accumulated future rewards obtained by executing them. 
To account for the increasing uncertainties associated with future rewards, the reward $r_{t+k}$ received $k$ time steps in the future is weighted by $\gamma^k$, where $\gamma\in[0,1]$.
In addition, we adopt the common approach of only including a finite number $T>1$ of future rewards in the return~\citep{sutton:98}, as follows:
\begin{equation}\label{eq:return}
  \begin{aligned}
    \mathcal R (\mathbf{s}_t,\pi) & = \sum_{k=0}^{T-1}\gamma^kr(\mathbf{s}_{t+k},\pi(\mathbf{s}_{t+k}),\mathbf{s}_{t+k+1}), \\
    \textnormal{with}\quad \mathbf{s}_{t+k+1} & = g(\mathbf{s}_{t+k},\mathbf{a}_{t+k}).
  \end{aligned}
\end{equation}
The overall state-independent policy performance $\mathcal{F}(\pi)$ is obtained by averaging $\mathcal R (\mathbf{s}_t,\pi)$ over all starting states $s_t \in S \subset \mathcal S$.
Thus, the optimal solutions to the RL problem are the policies $\hat{\pi}$ where
\begin{equation}\label{eq:fitness_function}
    \hat{\pi} \in \argmax_{\pi \in \Pi}\mathcal{F}(\pi), \quad\text{with}\quad \mathcal{F}(\pi)=\frac{1}{\abs{S}}\sum_{\mathbf{s}_t\in S}\mathcal R (\mathbf{s}_t,\pi).
\end{equation}
In optimization terminology, the policy performance function $\mathcal{F}(\pi)$ is known as the fitness function.

For most real-world industrial control problems, the cost of executing a potentially bad policy is prohibitive. 
Therefore, in model-based RL~\citep{busoniu:10}, the state transition function $g$ is approximated by a model $\tilde g$, that is either a first-principles model or has been created from previously recorded data. 
Substituting $\tilde{g}$ for the real system $g$ in \eqref{eq:return} allows us to obtain a model-based approximation $\tilde{\mathcal{F}}(\pi)$ of the true fitness function \eqref{eq:fitness_function}. 
Here, we consider models based on neural networks (NNs), but the proposed method could be extended to other models, such as Bayesian NNs~\citep{depeweg:16} and Gaussian process models~\citep{rasmussen:06}.

The model-based RL approaches considered in this paper are based on data sets $\mathcal{D}$ of state transition samples gathered from a real system. 
These samples are tuples $(\mathbf{s}_t,\mathbf{a}_t,\mathbf{s}_{t+1},r_{t+1})$ that represent a start state $\mathbf{s}_t$ transitioning to a next state $\mathbf{s}_{t+1}$ owing to take action $\mathbf{a}_t$ and yielding a reward $r_{t+1}$.
The set $\mathcal{D}$ can be generated using any policy (even a random one) prior to policy training and is subsequently used to generate world models $\tilde g$ that take inputs $(\mathbf{s}_t,\mathbf{a}_t)$ and predict $\mathbf{s}_{t+1}$ and $r_{t+1}$. 

\subsection{Fuzzy Controller}
\label{section:fuzzy}

Fuzzy set theory was first proposed by \citet{zadeh:65}. 
Based on this theory, \citet{mamdani:75} subsequently introduced so-called fuzzy controllers, specified by sets of linguistic if-then rules, whose membership functions can be activated independently to produce a combined output, computed by a suitable defuzzification function.

In a $D$-inputs-single-output system with $C$ rules, the fuzzy rules $R^{(i)}$ can be expressed as follows:
\begin{equation}
R^{(i)}: \text{ IF }\mathbf{s}\text{ is } m^{(i)} \text{ THEN }o^{(i)}, \quad \text{with }i\in\{1,\dotsc,C\}, 
\end{equation}
where $\mathbf{s}\in \mathbb{R}^{D}$ is the input vector (environment state, in our case), $m^{(i)}$ is the membership of a fuzzy set of the input vector in the premise part, and $o^{(i)}$ is a real number in the consequent part.

We use Gaussian membership functions~\citep{wang:92}.
These multivariate Gaussian functions are formed from products over all membership dimensions and yield smooth outputs, are local, and never produce zero activation.
We define each rule's membership function as follows:
\begin{align}
m^{(i)}(\mathbf{s})=\text{m}[\mathbf{c}^{(i)},\mathbf{\sigma}^{(i)}](\mathbf{s})&=\prod^{D}_{j=1}d(c_{j}^{(i)},\sigma_{j}^{(i)},s_j),\\
\text{with}\quad d(c,\sigma,s)&=\exp\left\{-\frac{(c-s)^2}{2\sigma^2}\right\},
\label{gaussian_membership}
\end{align}
where $m^{(i)}$ is the i-th parameterized Gaussian $\text{m}(\mathbf{c},\mathbf{\sigma})$, with center $\mathbf{c}^{(i)}$ and width $\mathbf{\sigma}^{(i)}$.

The output is determined by the following equation: 
\begin{equation}
\pi(\mathbf{s})=\tanh\left(\alpha\cdot\frac{\sum_{i=1}^{C}m^{(i)}(\mathbf{s})\cdot o^{(i)}}{\sum_{i=1}^{C}m^{(i)}(\mathbf{s})}\right),
\label{eq:defuzzifier}
\end{equation}
where the hyperbolic tangent limits the output to be between -1 and 1, and the parameter $\alpha$ can be used to change the function's slope.

\subsection{Fuzzy Particle Swarm Reinforcement Learning}
\label{section:fuzzy_pso}

FPSRL is a PSO-based approach to solve model-based batch RL problems~\citep{hein:17c}.
PSO is a population-based, non-convex, stochastic optimization heuristic and can be applied to any search space that is a bounded sub-space of a finite-dimensional vector space~\citep{kennedy:95}.

The position of each particle in the swarm represents a potential solution to the given problem. 
The particles \textit{fly} iteratively through the multidimensional search space, which is referred to as the fitness landscape. 
At each iteration, the particles move and receive fitness values for their new positions.
These values are used to update each particle's velocity vector as well as those of all the other particles in a certain neighborhood.

For a given maximization problem, the best particle positions at iteration $p$ are calculated as follows:
\begin{equation}
\label{personal_best}
\mathbf{y}_i(p)=\begin{cases}
\mathbf{x}_i(p), & \text{if }\mathcal{F}(\mathbf{x}_i (p))>\mathcal{F}(\mathbf{y}_i (p-1))\\
\mathbf{y}_i(p-1), & \text{else},
\end{cases}
\end{equation}
where, in our framework, $\mathcal{F}$ is the fitness function given in \eqref{eq:fitness_function} and the particle positions represent the policy parameters $\mathbf{x}$.

The parameter vector $\mathbf{x}\in\mathcal X$, where $\mathcal X$ is the set of valid Gaussian fuzzy parameterizations, is of size $(2D+1)\cdot C + 1$ and can be presented as follows:
\begin{align}
\label{eq:x_vector}
\begin{split}
\mathbf{x}=( & c_{1}^{(1)},c_{2}^{(1)},\dotsc,c_{D}^{(1)},\sigma_{1}^{(1)},\sigma_{2}^{(1)},\dotsc,\sigma_{D}^{(1)},o^{(1)},\\
& c_{1}^{(2)},c_{2}^{(2)},\dotsc,c_{D}^{(2)},\sigma_{1}^{(2)},\sigma_{2}^{(2)},\dotsc,\sigma_{D}^{(2)},o^{(2)},\dotsc,\\
& c_{1}^{(C)},c_{2}^{(C)},\dotsc,c_{D}^{(C)},\sigma_{1}^{(C)},\sigma_{2}^{(C)},\dotsc,\sigma_{D}^{(C)},o^{(C)},\alpha).
\end{split}
\end{align}

\subsubsection{Rule Construction}

To set up the FPSRL training process, we must take an assumption about the number of rules per policy during the rule construction step (Fig.~\ref{figure:compare}).
This requires either prior knowledge about the current problem, or testing different numbers of rules combinatorially~\citep{hein:17c}.

\subsubsection{Feature Selection}

Industrial applications usually have dozens or even hundreds of possible state features, collected by different plant sensors.
However, in our experience, very often only a small subset of them are required to create a policy that performs well.
Unfortunately, determining the most important features is an expensive and ambiguous process, but it is nonetheless essential if we are to apply promising techniques such as FPSRL to industrial applications.
Therefore, we propose a two-step approach that yields lists of features for each action, which are ordered by their relevance to an optimal policy.

First, an optimal trajectory is generated by applying the PSO-P receding horizon controller~\citep{hein:16,hein:18} to the system model $\tilde{g}$.
Here, PSO-P uses the model $\tilde{g}$ to determine action sequences $(\mathbf{a}_t,\mathbf{a}_{t+1},\ldots,\mathbf{a}_{T})$ starting from the state $\mathbf{s}_t$.
The first action $\mathbf{a}_t$ in the optimal sequence is stored in the tuple $(\mathbf{s}_t,\mathbf{a}_t)$, and this process is repeated for all possible states in the data set $\mathcal{D}$.
Note that no explicit policy representation is required to use the model to generate optimal actions.

Second, the AMIFS feature selection heuristic~\citep{tesmer:04} is used to order the possible features $(s_1,s_2,\ldots)=\mathbf{s}$ of state $\mathbf{s}$ in terms of their relevance to the individual action dimensions $(a_1,a_2,\ldots)=\mathbf{a}$.
AMIFS uses mutual information as a measure of feature relevance and redundancy to reveal non-linear feature-action relations.

Finally, the resulting ordered feature lists, e.g., $\{s_3,s_2,s_1\}_{a_1}$ and $\{s_2,s_1,s_3\}_{a_2}$ for the action dimensions $a_1$ and $a_2$, are used to construct compact and interpretable fuzzy rule representations, whose parameters will then be optimized by FPSRL.

\subsection{Fuzzy Genetic Programming Reinforcement Learning}
\label{section:fuzzy_gp}

In this section, we introduce a GP-based approach that can generate interpretable fuzzy rule policies automatically using model-based batch RL.
Analogously to FPSRL, FGPRL uses an approximate system model to predict the performance of policy candidates and subsequently uses this knowledge to iteratively generate high-performing policies.
Unlike FPSRL, FGPRL not only optimizes the predefined policy parameters but also automatically selects the relevant state features, finds the necessary number of rules, and returns a Pareto front of policy candidates with different levels of complexity. 

\begin{figure}
	\centering
	\includegraphics[width=\individualwidth]{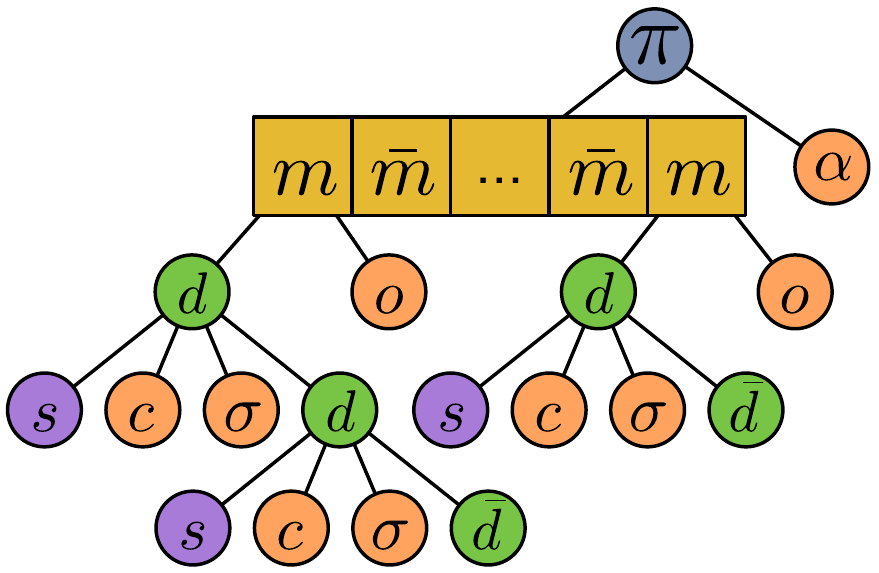}
	\caption{A fuzzy GP individual}
	\label{gp_individual}
\end{figure}

FGPRL is based on GP, which encodes computer programs as sets of genes and then modifies (evolves) them using a so-called genetic algorithm (GA) to drive the optimization of the population.
The solution spaces comprise computer programs that perform well on the given tasks~\citep{koza:92}.
Since we are interested in using interpretable fuzzy controllers as RL policies, the genes in our case include membership and defuzzification functions, as well as constant floating-point numbers and state variables. 
These fuzzy policies can be represented as function trees (Fig.~\ref{gp_individual}) and stored efficiently in arrays.

The GA drives the optimization process using selection and reproduction of population members, both of which are based on the members' fitness values $\mathcal{F}$.
These represent how well each individual can perform the given task.
Selection ensures that only the fittest individuals will survive into the next generation.
Similar to the case of biological breeding process, pairs of individuals are selected for reproduction based on their fitness, and two offspring individuals are created for each pair by crossing their chromosomes.
Technically, this is achieved by selecting compatible cutting points in the function trees and interchanging the subtrees below these cuts.
Here, we apply tournament selection~\citep{blickle:95} to select the parent individuals.
In addition, we use strongly-typed GP for FGPRL to avoid constructing ill-defined rules~\citep{alba:99}.
This means that each building block is assigned a type (Table~\ref{table:types_complexities}).
The different colors in Fig.~\ref{gp_individual} highlight the example GP individual's type structure.
During the crossover process, only cutting points of equal type (color) are selected to ensure that only legal offspring are created.

\begin{table}
	\begin{center}
		\begin{tabular}{llclr} 
			\addlinespace[-\aboverulesep] 
 			\cmidrule[\heavyrulewidth]{1-2}\cmidrule[\heavyrulewidth]{4-5}
 			\multicolumn{2}{c}{Type} & & \multicolumn{2}{c}{Complexity}\\
 			\cmidrule{1-2}\cmidrule{4-5}
			Variable & $s$ & &  $\pi,\bar{m},\bar{d}$ & 0\\
			Floating-point number & $c,\sigma, o,\alpha$ & & $s$ & 1\\
			Dimension & $d,\bar{d}$ & & $c,\sigma, o,\alpha$ & 1\\
			Rule & $m,\bar{m}$ & & $d$ & 2\\
			Policy & $\pi$ & & $m$ & 10\\
 			\cmidrule[\heavyrulewidth]{1-2}\cmidrule[\heavyrulewidth]{4-5}
			\addlinespace[-\belowrulesep]
		\end{tabular}
		\caption{Types and complexities of the GP building blocks.
		The terminal blocks $\bar{m}$ and $\bar{d}$ have values 0 and 1, respectively.}
		\label{table:types_complexities}
	\end{center}
\end{table}

We adopt the so-called Gaussian mutator as the mutation operator for floating-point terminals, which is common for evolutionary algorithms~\citep{schwefel:81,schwefel:95}.
In each generation, a given fraction of the best-performing individuals is selected for each level of rule complexity.
Then, these individuals are copied, and their original floating-point values $z$ are mutated by drawing replacement values $z'$ from the normal distribution $\mathcal{N}(z,0.1\abs{z})$.
If the best copy's performance is better than that of the original individual, it is added to the new population.
This allows us to conduct a local search in the policy space because it does not affect the individual's basic genotype structure.

To yield fuzzy rules with the structure described in Section~\ref{section:fuzzy}, we must apply an additional tree correction.
The GA can construct rules where two or more activation functions act on the same state variable, such as in Fig.~\ref{gp_individual} where the same $s$ appears twice below the same rule $m$.
Such rules are expected to be difficult to interpret since their shape does not conform the standard Takagi-Sugeno fuzzy inference model.
For FGPRL, we decided to check every tree before evaluating its fitness by looking for recurring state variables and cutting out their corresponding activation functions.
Note that the structures of the subsequent activation functions are not affected.

Since we are looking for interpretable solutions, we need to establish a suitable complexity measure.
An individual's complexity can generally be measured in terms of its genotype (structural) or phenotype (functional)~\citep{le:16}.
Here, we decided to use a simple node counting measurement strategy where different types of functions, variables, and terminals were weighted differently.
Table~\ref{table:types_complexities} lists the weights (complexities) we decided to use in our experiments.
Note that the weights yield a problem-specific balance between learning controllers consisting of more rules with less dimensions and vice versa.

Finally, we decided to create new generations for FGPRL using the following ratios: 45\% crossover, 5\% reproduction, 10\% mutation, and 40\% new random individuals.

\subsubsection{Local Search}

Since FGPRL's GP process searches the entire fuzzy policy structure space, it is prone to underestimate the importance of local fuzzy parameter tuning~\citep{moscato:89,tsakonas:13}.
We propose to counteract this by applying an additional parameter tuning step to all the terminals ($c,\sigma,o,\alpha$) after optimization is complete (Fig.~\ref{figure:compare}).
Applying PSO to every individual in the final Pareto front yields an updated front comprising at most the same number of individuals with equal or higher fitness values.

\section{Experiments}
\label{section:experiments}

In this section, we evaluate both approaches, FPSRL and FGPRL, using two benchmarks.
The first is the so-called cart-pole swing-up problem (CP), a widely known RL benchmark, that was selected to demonstrate the methods' performance on a task where they could be easily compared with other RL methods.
However, CP has a low-dimensional state space and its dynamics are deterministic and smooth.
To investigate these methods' performance on real-world industry applications, we also selected a second benchmark, the industrial benchmark (IB).
This combines a high-dimensional state space with stochastic dynamics, thus making its results increasingly meaningful for industrial applications, such as controlling wind and gas turbines.

To compare the computational costs of FPSRL with FGPRL, we decided to use the parameter values shown in Table~\ref{table:calculation} for our experiments.
Unlike FPSRL, FGPRL produces a whole Pareto front of solutions with different complexity and fitness values, while FPSRL only yields one solution, derived from its initial rule set.
Note that although it is useful for FGPRL, FPSRL does not require any additional local optimization.
Consequently, we chose the \textit{additional local search} and \textit{runs for multiple complexities} settings to produce similar \textit{total fitness value calculations}, yielding the results presented below.

\begin{table}
	\centering
		\begin{tabular}{lcrr}
			\toprule
			& & FPSRL & FGPRL\\
			\cmidrule[\heavyrulewidth]{1-1}\cmidrule[\heavyrulewidth]{3-4}
			Particles/Individuals & & $1,000$ & $1,000$\\
			Iterations/Generations & & $\times\ 10,000$ & $\times\ 10,000$\\
			\cmidrule{1-1}\cmidrule{3-4}
			Fitness value calculations (A) & & $1\mathrm{e}{+7}$ & $1\mathrm{e}{+7}$\\
			\cmidrule[\heavyrulewidth]{1-1}\cmidrule[\heavyrulewidth]{3-4}
			Optimization result & & Single policy & Pareto front of policies\\
			& & 1 & 22 to 41\\
			Additional local search & & $\times\ 0$ & $\times\ 1\mathrm{e}{+6}$\\
			\cmidrule{1-1}\cmidrule{3-4}
			Additional local fitness value calc. (B) & & $0$ & $2.2\mathrm{e}{+7}$ to $4.1\mathrm{e}{+7}$\\
			\cmidrule[\heavyrulewidth]{1-1}\cmidrule[\heavyrulewidth]{3-4}
			Runs for multiple complexities & & 4 & 1\\
			(A) & & $\times\ 1\mathrm{e}{+7}$ & $\times\ 1\mathrm{e}{+7}$\\
			\cmidrule{1-1}\cmidrule{3-4}
			Fitness value calc. for multiple compl. (C) & & $4\mathrm{e}{+7}$ & $1\mathrm{e}{+7}$\\
			\cmidrule[\heavyrulewidth]{1-1}\cmidrule[\heavyrulewidth]{3-4}
			Total fitness value calc. (B)+(C) & & $4\mathrm{e}{+7}$ & $3.2\mathrm{e}{+7}$ to $5.1\mathrm{e}{+7}$\\
			\bottomrule	
		\end{tabular}
	\caption{Computational costs for FPSRL and FGPRL}
	\label{table:calculation}
\end{table}

\subsection{Cart-pole Swing-up}

\subsubsection{Dynamics}

The objective of the CP benchmark is to apply forces to a cart moving along a one-dimensional track to keep a pole (hinged to the cart) in an upright position. 
The four Markov state variables are the pole's angle $\theta$ and angular velocity $\dot\theta$, and the cart's position $\rho$ and velocity $\dot\rho$. 
These variables completely describe the Markov state; therefore, no additional information is required about the system's previous behavior. 
The RL agent's task is to find a sequence of force actions $a_t,a_{t+1},a_{t+2},\ldots$ that prevent the pole from falling over~\citep{fantoni:02}.
The CP experiments described in this paper were conducted using the $CLS^2$ software\footnote{http://ml.informatik.uni-freiburg.de/research/clsquare.}.

There are no restrictions on the cart's position or the pole's angle. 
Consequently, the pole can swing through, which is an important property of the CP. 
Since the pole's angle can initially be anywhere in the full interval $[-\pi,\pi]$, it is often necessary for the policy to swing the pole from one side to another to gain sufficient momentum to raise it and consequently receive the highest reward.

CP policies can apply actions of between $-30$~N and $+30$~N to the cart, and the reward function is given as follows: 
\begin{equation}
	r(\mathbf{s})=
	\begin{cases}
		0, & \text{if }\abs{\theta'}<0.5 \text{ and } \abs{\rho'}<0.5,\\
		-1, & \text{otherwise}.
	\end{cases}
\end{equation}

\subsubsection{Benchmark Setup}

Initially, we generated data set $\mathcal{D}$ by applying random actions using the real CP dynamics.
Generating 100 state-action trajectories of length 100 gave us $\abs{\mathcal{D}}=10,000$.
These trajectories' initial states $(\theta,\dot{\theta},\rho,\dot{\rho})$ were sampled uniformly from $[-\pi, \pi]\times\{0\}\times\{0\}\times\{0\}$.
Then, we trained five NN system models $\tilde{g}$, one for each state variable and one for predicting the probability of reaching the goal region~\citep{hein:17c}.

These system models $\tilde{g}$ were subsequently used in model-based RL, with a time horizon $T$ of 500 and a discount factor $\gamma$ of 0.994.
The training process involved 100 training states sampled from $[-\pi, \pi]\times\{0\}\times[-0.5, 0.5]\times\{0\}$.
Solutions with $\mathcal{F}\approx-50$ or greater were considered to be successful because such policies could swing-up more than 99\% of the given test states.

Finally, the best policies were tested against the real system dynamics using the same $T$ and $\gamma$ parameter values but a different set of 100 test states sampled from $[-\pi, \pi]\times\{0\}\times[-0.5, 0.5]\times\{0\}$.

\subsubsection{Results}

Here, we compare the results of CP experiments in which we ran the benchmark for each method 10 times.
FPSRL produced 40 policies for 4 complexity levels, while FGPRL produced 278 policies for 96 complexity levels.
Note that both methods involved a similar number of fitness value calculation (Table~\ref{table:calculation}).

Since it is known that, for the CP, all four Markov state variables are required to produce policies that perform well, we skipped the feature selection step for FPSRL (Fig.~\ref{figure:compare}) and invested the fitness value calculations budget in  evaluating different numbers of rules, i.e., 2, 4, 6, and 8 rules yielding complexities of 63, 125, 187, and 249, respectively.

\begin{figure}
	\centering
	\includegraphics[width=\cpparetowidth]{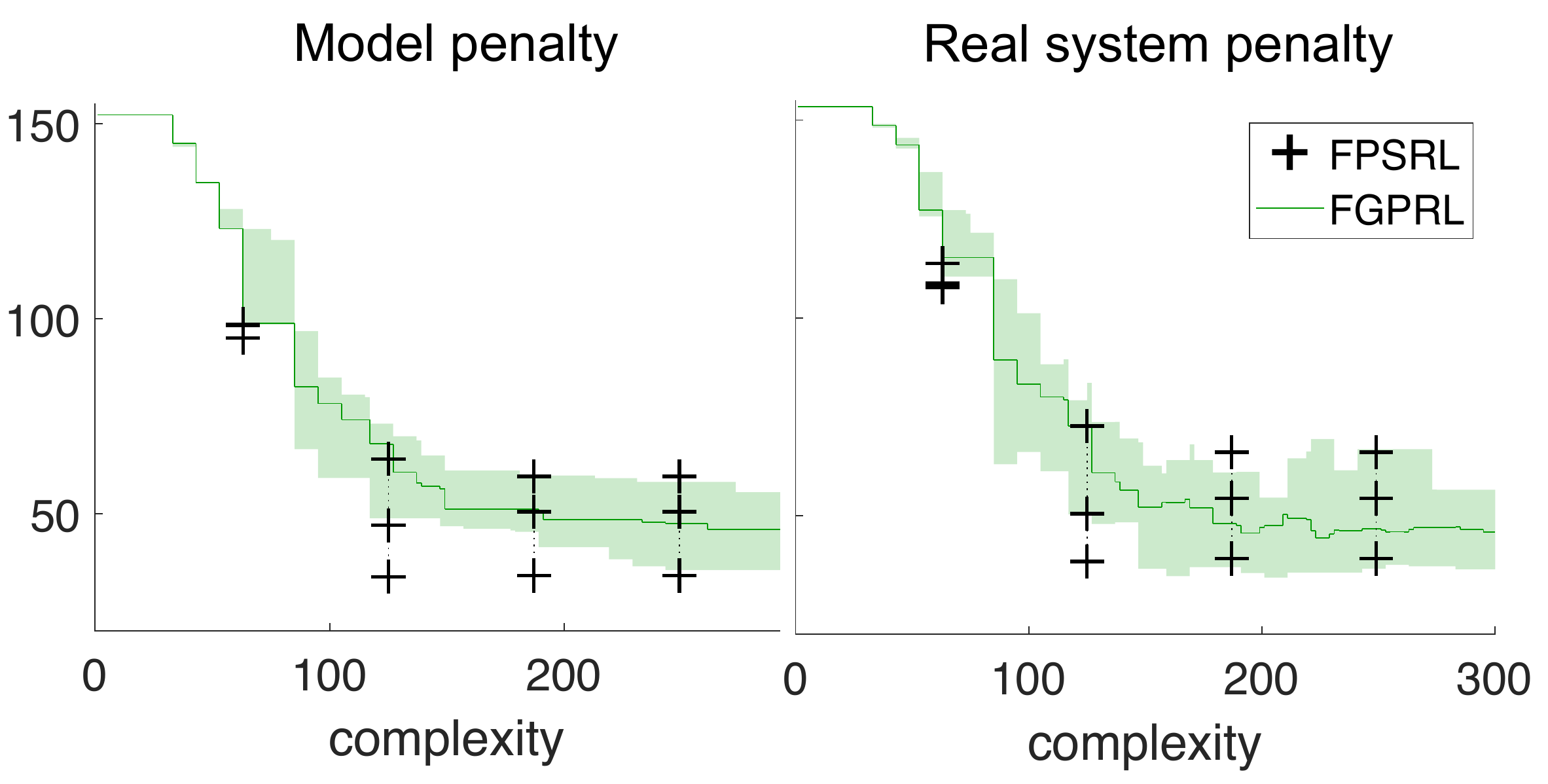}
	\caption{Comparison of FPSRL and FGPRL performance for the CP, for both the model and the real system.
		Here, the penalty is equal to the negative fitness.
		The green lines represent the median Pareto fronts over the FGPRL experiments.
		The semi-transparent green areas show the minimum and maximum penalties obtained from the experiments, while the black crosses indicate the FPSRL results.
		These show, from top to bottom, the maximum, median, and minimum penalties for each complexity level.
		The Pareto fronts resulting from training on the system model are on the left, while the right-hand plot shows the manner in which testing the same policies against the real system dynamics \textit{altered} the fronts.}
	\label{figure:cp_pareto_small}
\end{figure}

Fig.~\ref{figure:cp_pareto_small} shows that for problems such as the CP, which have low-dimensional state spaces and no irrelevant or redundant state variables, applying prior knowledge to the rule construction step yields a rule structure that can be easily tuned to produce high-performance, interpretable fuzzy policies.
FPSRL can utilize all the available computational resources to tune a fixed set of parameters.

However, FGPRL has to employ the same resources to search a significantly larger space of possible solutions.
Note that although FGPRL is theoretically able to produce exactly the same fuzzy policies for a complexity of 63 as FPSRL, it was unable to find a comparable solution for the system models in any of our experiments (Fig.~\ref{figure:cp_pareto_small}).
The best individual it produced with a complexity of 63 or less had a penalty value of 48.88, which was even above the median FPSRL penalty value of 47.05.

Comparing the model and real dynamics penalties yields another interesting observation.
Even though the best FGPRL policies never surpassed FPSRL's performance for complexities of 300 and below on the system model, when they were evaluated using the real dynamics, some FGPRL policies actually performed better than the FPSRL policies.
This could possibly be because the swarm optimization had already started to overfit the fuzzy parameters with respect to the system model; this means that FPSRL was exploiting model inaccuracies, thus reducing its performance on the real dynamics.

\subsection{Industrial Benchmark}

\subsubsection{Dynamics}

The IB\footnote{\url{http://github.com/siemens/industrialbenchmark}} was designed to simulate several of the common challenges associated with many industrial applications~\citep{hein:17a}. 
It was not designed to approximate any specific real-world system but to be of a comparable hardness and complexity to many industrial applications.

The IB's state space is continuous, high-dimensional, and only partially observable. 
The actions are made up of three continuous components and affect three control inputs. 
In addition, the IB includes stochastic and delayed effects. 
The optimization task also involves multiple criteria; there are two reward components with opposite dependencies on the actions. 
Its dynamics are heteroscedastic, with state-dependent observation noise and probability distributions that are based on latent variables. 
Finally, it depends on an external driver that cannot be influenced by the actions.

At any given time step $t$, the RL agent can influence the state via actions $\mathbf{a}_t\in[-1,1]^3$ that change the three observable state control variables, namely the velocity $v$, gain $g$, and shift $h$, i.e., $\mathbf{a}_t = \left(\Delta v_t,\Delta g_t, \Delta h_t\right)$.

The state $\mathbf{s}_t$ and successor state $\mathbf{s}_{t+1}$ are the Markov environment states.
They can only be partially observed by the agent. 
In addition to the three control variables, $v$, $g$, and $h$, there is a setpoint $p$ that simulates an external force, such as the  load on a power plant or the speed of the wind driving a turbine that the agent cannot control but still has a significant influence on the system's dynamics. 
The system also suffers from a detrimental fatigue $f_t$, which depends on the setpoint $p$ and the chosen control values $\mathbf{a}_t$, and consumes resources, such as power and fuel, represented by the consumption $c_t$. 
At each time step, it generates output values for $c_{t+1}$ and $f_{t+1}$, which are part of the internal state $\mathbf{s}_{t+1}$.
The reward is calculated as $r_{t+1}=-c_{t+1}-3f_{t+1}$.

Note that the IB system's complete Markov state $\mathbf{s}$ is unobservable. 
Only the observation vector $\mathbf{o}=(v,g,h,p,c,f)\subset\mathbf{s}$ can be observed externally.
The Markov state $\mathbf{s}_{t}$ can be approximated using a sufficient number of historic observations $\left(\,\mathbf{o}_{t-H},\mathbf{o}_{t-H+1},\ldots,\mathbf{o}_{t}\right)$ with a time horizon $H$. 
A system model $\tilde{g}\left(\,\mathbf{o}_{t-H},\mathbf{o}_{t-H+1},\ldots,\mathbf{o}_{t},\mathbf{a}_t\right)$ that computes $\left(\mathbf{o}_{t+1},r_{t+1}\right)$ with $H=30$, achieved adequate prediction performance during our IB experiments. 
Note that combining the six-dimensional observation vector with a time horizon $H$ of 30 results in a 180-dimensional approximate Markovian state vector.

\subsubsection{Benchmark Setup}

The system was initialized for each setpoint $p$ in $\{10,20,\ldots,100\}$ and then random trajectories of length 1,000 were produced.
This process was repeated 10 times, resulting in a data set of size $\abs{\mathcal{D}}=100,000$.
Following the approach reported in \citep{hein:17b}, we trained two recurrent NNs $\tilde{g}$ to predict the consumption $c$ and fatigue $f$.
These models $\tilde{g}$ were then used in model-based RL, with a time horizon $T$ of 100 and a discount factor $\gamma$ of 0.97.
The training process involved 100 training states, drawn randomly from states in $\mathcal{D}$.

Finally, the best policies were tested against the real system dynamics, using the same $T$ and $\gamma$ parameter values, but a different set of 100 test states drawn randomly from states in $\mathcal{D}$.

\subsubsection{Results}

As with the CP experiments, we ran the IB benchmark 10 times for each method.
FPSRL produced 40 policies for 4 complexity levels, while FGPRL produced 368 policies for 86 complexity levels.

To construct interpretable fuzzy rules using FPSRL, we have to select suitable features before swarm optimization of the fuzzy parameters (Section~\ref{section:fuzzy_pso}).
The proposed method identified the following state variables as being the most important for each action dimension: $\Delta v$: $\{f_0,v_7,c_13,h_0\}$, $\Delta g$: $\{c_0,f_1,c_{15},p\}$, and $\Delta h$: $\{h_4,p,h_{10},h_0\}$.
Here, the variables are listed in descending order of importance and the indices represent the time elapsed since the observation.
Four different fuzzy rule structures were constructed based on these variables.
The first policy with a complexity of 99, incorporated only the first variable in each list into two rules per action dimension, while the other policies, with complexities of 129, 159, and 189, incorporated the first two, the first three, or all four variables, respectively.

\begin{figure}
	\centering
	\includegraphics[width=\ibparetowidth]{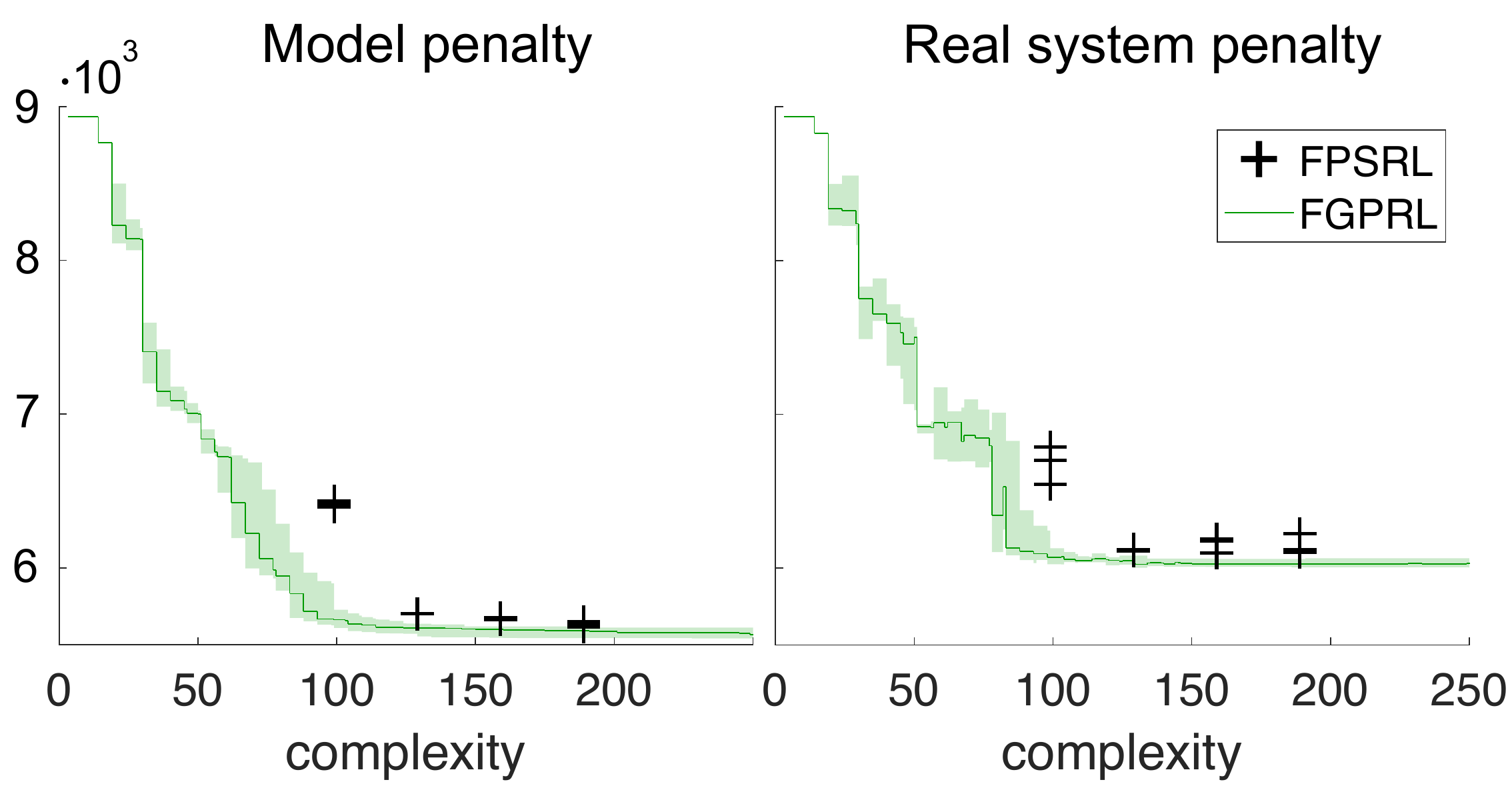}
	\caption{Comparison of FPSRL and FGPRL performance for the IB, for both the model and the real system}
	\label{figure:ib_pareto_small}
\end{figure}

\begin{figure}
\centering
\subfloat[FPSRL]{
	\includegraphics[trim=0 0 0 0,width=\rulewidth]{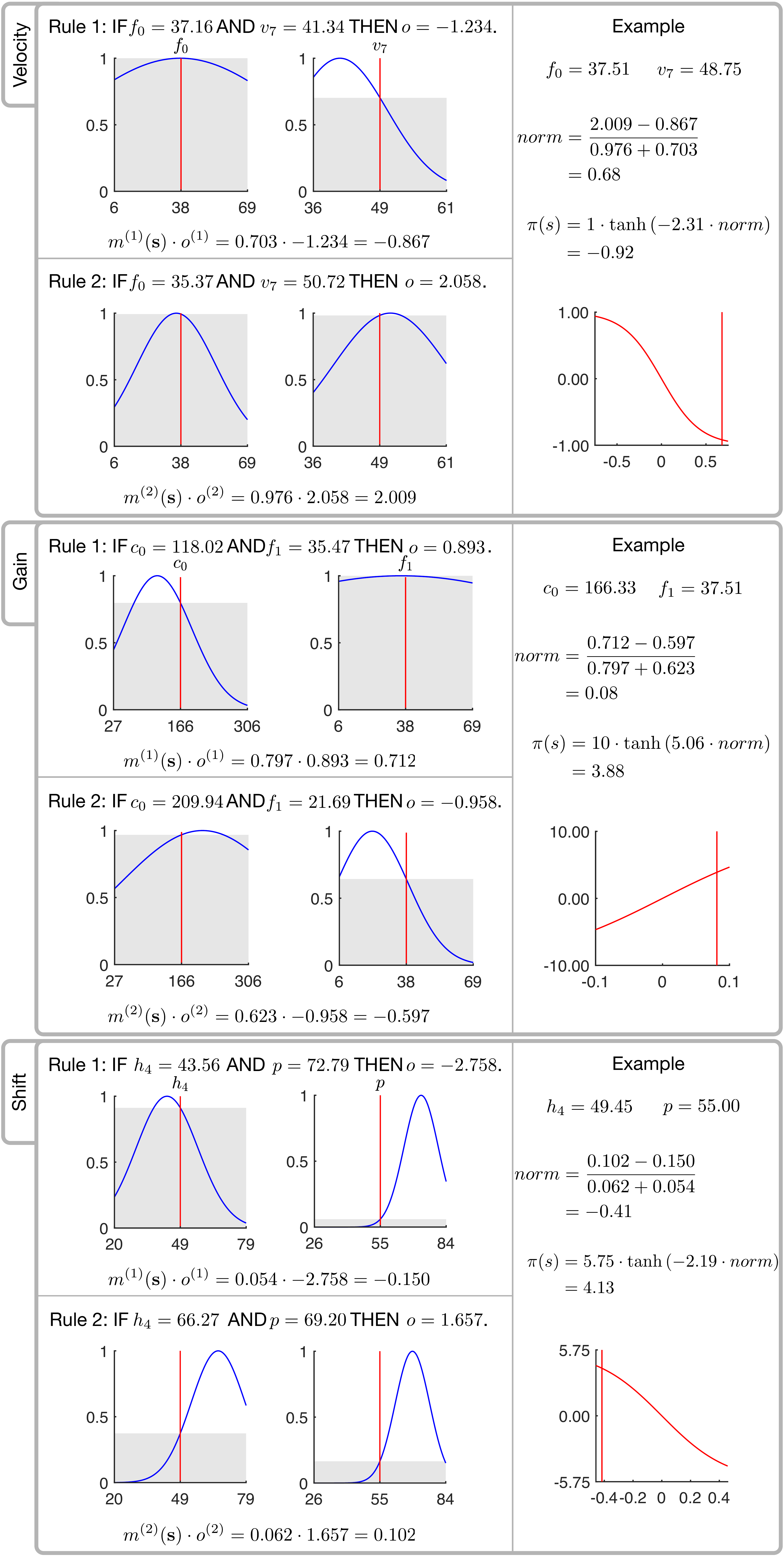}
	\label{figure:ib_rules_fpsrl}
}
\subfloat[FGPRL]{
	\includegraphics[trim=0 0 0 0,width=\rulewidth]{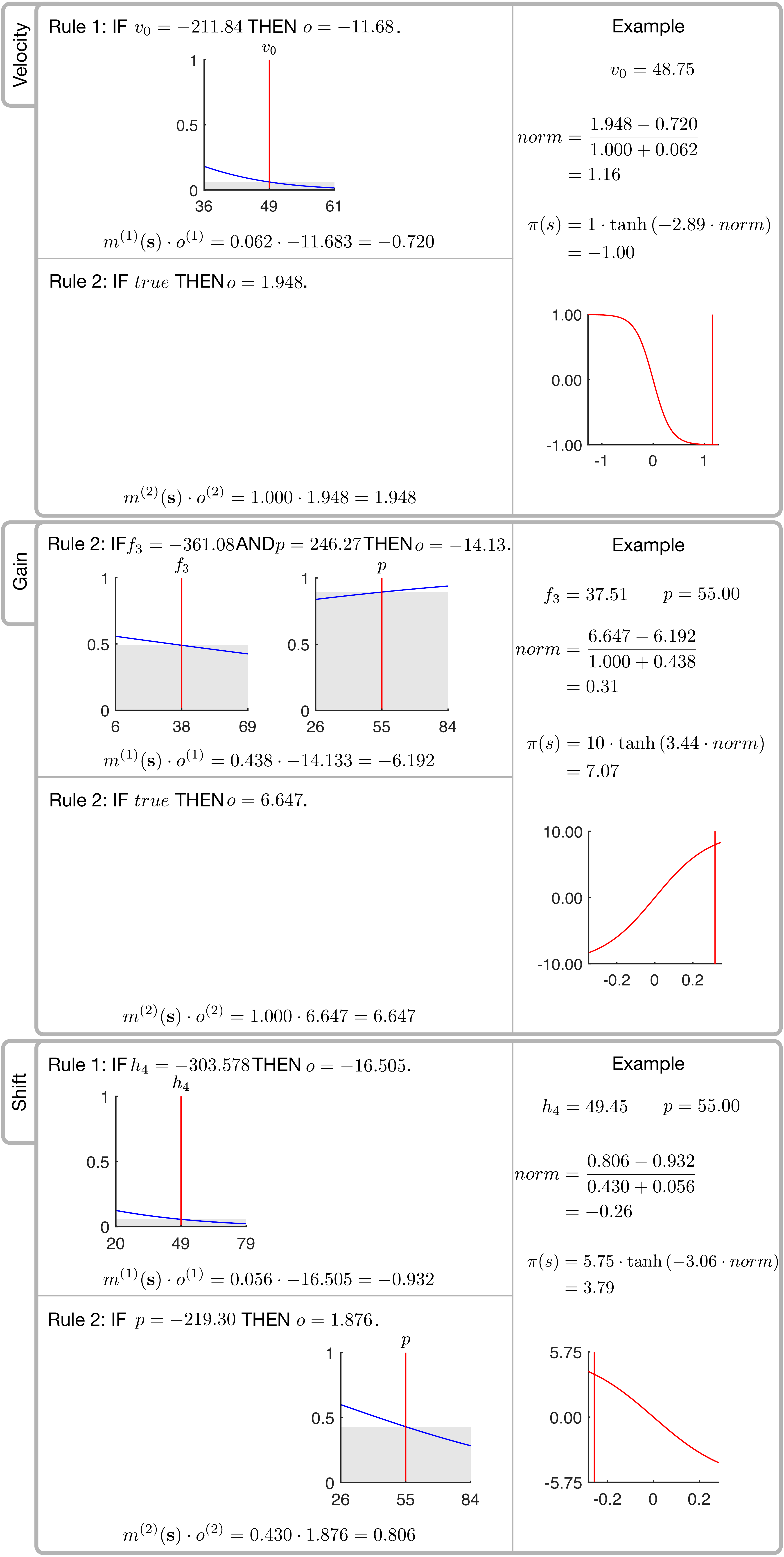}
	\label{figure:ib_rules_fgprl}
}
\caption{Comparison of the fuzzy policies produced by FPSRL and FGPRL for the IB.
	Although both policies have similar fitness values (\ref{figure:ib_rules_fpsrl}: -5700, \ref{figure:ib_rules_fgprl}: -5635), their complexity is very different.
	The FGPRL policy only involves the minimum number of state variables required to yield adequate performance, meaning that the fuzzy controller is substantially more interpretable.}
\label{figure:ib_rules}
\end{figure}

Using the proposed feature selection heuristic, FPSRL was able to generate policies with adequate performance ($\tilde{\mathcal{F}}=-5700$) for complexities of 129 or higher (Fig.~\ref{figure:ib_pareto_small}).
However, FGPRL was able to generate policies of a significant low complexity with a higher performance, achieving $\tilde{\mathcal{F}}=-5635$ at a complexity of 94 (Fig.~\ref{figure:ib_rules}).
Moreover, the FGPRL search space covers all possible combinations of state dimensions and numbers of rules for each individual action.
For industrial problems where the state-to-action dependencies are not known a priori, we expect this ability to search autonomously to be highly valuable to control system designers and domain experts.

Comparing the policies' performance on the approximate IB model with their performance on the real IB dynamics shows that the results of training can be transferred to the real system as long as their regression and generalization quality is adequate (Fig.~\ref{figure:ib_pareto_small}).

\section{Conclusion}

In this paper, we have evaluated two approaches to learn fuzzy control policies autonomously in terms of their performance and interpretability in industrial applications.
We  have considered applications with high-dimensional continuous state and action spaces and have proposed a feature selection heuristic that enables the previously presented FPSRL approach to be applied successfully in such industrial domains.
Our second contribution is a GP-based fuzzy policy learning approach called FGPRL that utilizes the same model-based batch RL technique as FPSRL.
However, instead of only tuning the parameters of fixed fuzzy policy structures, FGPRL searches the full space of all possible fuzzy controllers by determining the important state variables and the number of rules required and subsequently tunes all the rule parameters.

Experiments using the standard CP RL benchmark showed that FPSRL has a significant advantage when no feature selection is necessary and the number of rules required can be easily determined by simply testing a few different options.
In this case, FGPRL's significantly wide search space is a drawback and it was far less likely to converge to a solution with similar performance to that produced by FPSRL when using a similar number of fitness value calculations.
However, FGPRL was occasionally able to produce high-performance solutions for the CP benchmark.

Experiments using the IB, a benchmark that mimics real industrial systems like gas or wind turbines, yielded significant advantage for FGPRL over FPSRL.
This benchmark has a high-dimensional state space, a multidimensional action space, stochastic and delayed effects, and a reward function with multiple criteria.
Applying feature selection to FPSRL and manually testing different fuzzy policy structures did not yield satisfactory performance for low complexity solutions.
In contrast, FGPRL was able to find high-quality interpretable solutions of low complexity with a similar number of fitness value calculations.

These results indicate that FGPRL is better than FPSRL at creating interpretable fuzzy policies autonomously from existing transition samples.

\section*{Acknowledgments}

The project this report is based on was supported with funds from the German Federal Ministry of Education and Research under project number 01IB15001. 
The sole responsibility for the report's contents lies with the authors.

\bibliographystyle{ACM-Reference-Format}
\bibliography{bibliography} 

\end{document}